\documentclass{article}

\usepackage{microtype}
\usepackage{graphicx}
\usepackage{subfigure}
\usepackage{booktabs} 

\usepackage{hyperref}

\usepackage[accepted]{icml2025}

\usepackage{amsmath}
\usepackage{amssymb}
\usepackage{mathtools}
\usepackage{amsthm}

\usepackage[capitalize,noabbrev]{cleveref}

\theoremstyle{plain}

\theoremstyle{definition}

\theoremstyle{remark}

\usepackage[textsize=tiny]{todonotes}

\icmltitlerunning{\textsc{AutoCircuit-RL}: Reinforcement Learning-Driven LLM for Automated Circuit Topology Generation}

\begin{document}

\twocolumn[
\icmltitle{\textsc{AutoCircuit-RL}: Reinforcement Learning-Driven LLM \newline for Automated Circuit Topology Generation}



\icmlsetsymbol{equal}{*}

\begin{icmlauthorlist}
\icmlauthor{Prashanth Vijayaraghavan}{equal,comp}
\icmlauthor{Luyao Shi}{equal,comp}
\icmlauthor{Ehsan Degan}{comp}
\icmlauthor{Vandana Mukherjee}{comp}
\icmlauthor{Xin Zhang}{comp2}

\end{icmlauthorlist}

\icmlaffiliation{comp}{IBM Almaden Research Center, San Jose, CA 95120, USA}
\icmlaffiliation{comp2}{IBM Thomas J. Watson Research Center, Yorktown Heights, NY 10598}

\icmlcorrespondingauthor{Prashanth Vijayaraghavan}{prashanthv@ibm.com}
\icmlcorrespondingauthor{Luyao Shi}{luyao.shi@ibm.com}
\icmlcorrespondingauthor{Xin zhang}{xzhang@us.ibm.com}

\icmlkeywords{Machine Learning, ICML}

\vskip 0.3in
]



\printAffiliationsAndNotice{\icmlEqualContribution} 


\begin{abstract}
Analog circuit topology synthesis is integral to Electronic Design Automation (EDA), enabling the automated creation of circuit structures tailored to specific design requirements. However, the vast design search space and strict constraint adherence make efficient synthesis challenging. Leveraging the versatility of Large Language Models (LLMs), we propose \textsc{AutoCircuit-RL}, a novel reinforcement learning (RL)-based framework for automated analog circuit synthesis. The framework operates in two phases: instruction tuning, where an LLM learns to generate circuit topologies from structured prompts encoding design constraints, and RL refinement, which further improves the instruction-tuned model using reward models that evaluate validity, efficiency, and output voltage. The refined model is then used directly to generate topologies that satisfy the design constraints. Empirical results show that \textsc{AutoCircuit-RL} generates $\sim$12\% more valid circuits and improves efficiency by $\sim$14\% compared to the best baselines, while reducing duplicate generation rates by $\sim$38\%. It achieves over 60\% success in synthesizing valid circuits with limited training data, demonstrating strong generalization. These findings highlight the framework's effectiveness in scaling to complex circuits while maintaining efficiency and constraint adherence, marking a significant advancement in AI-driven circuit design.
\end{abstract}

\section{Introduction}

AI and machine learning have been applied to various circuit design tasks, including parameter optimization~\cite{Sizing_MIT} and physical design~\cite{hakhamaneshi2019bagnet}, which focus on circuit optimization with a fixed circuit topology. Analog circuit topology synthesis~\cite{bengio2013estimating} is a fundamental aspect of EDA, where the configuration and interconnection of components directly influence circuit functionality and performance. Despite years of EDA advancements, automation of analog circuit topology synthesis has remained underexplored until recently.

\begin{figure}[!t]
\centering
\includegraphics[width=\linewidth]{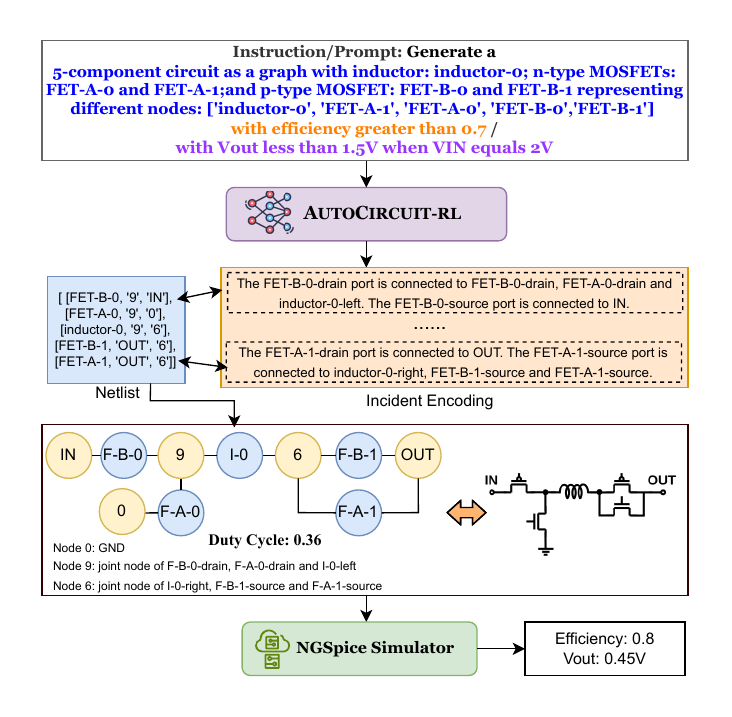}
\caption{Given a prompt with design constraints, the goal is to generate a circuit topology and duty cycle satisfying those constraints. Component constraints (blue) are mandatory, while efficiency (orange) and voltage (purple) constraints are optional. In the output, blue circles denote components (``F'' for FET, ``I'' for inductor), and yellow circles denote connection nodes.}

\label{fig:overview}
\end{figure}
The key challenge in circuit topology synthesis stems from the exponential growth of the design space with the number of components, making high-quality designs rare and hard to discover. This sparsity makes it difficult to satisfy specific performance and design constraints. Manual topology design remains time-consuming and requires significant expertise, while brute-force or random search methods are computationally infeasible due to the vastness of the space.

Existing AI methods fall into two categories: \textbf{(a) AI-based search-based algorithms}, including rule-based systems, heuristics, genetic algorithms~\cite{Topo_GA}, and tree-based search~\cite{ICCAD21, Topo_Tree}. While partially effective, these methods struggle with scalability, efficiency, and adaptability to evolving design requirements. They often require numerous queries and long runtimes to find circuits that meet targets. For example, Fan~\cite{ICCAD21} uses tree sampling for automated topology design but faces scalability and practical challenges when handling diverse performance needs, requiring over 400 simulation queries per design. Similarly, other search-based methods~\cite{Topo_Tree} demand extensive simulations for new specifications. \textbf{(b) Generative AI-based frameworks}, including graph-based and LLM-based generative methods. Graph generative models use VAEs to produce netlists as undirected~\cite{simonovsky2018graphvae} or directed graphs~\cite{dong2023cktgnn, zhang2019d}, but lack precise control over component count, efficiency, or power conversion ratio. Recently, LLMs have been applied to automated circuit topology synthesis \cite{vijayaraghavan2024circuitsynth, icml2024, lai2024analogcoder}, leveraging their pattern learning and design generation abilities. Unlike search-based methods, LLMs produce circuits from a single prompt after training, enabling faster generation.

However, most LLM approaches are limited in scale and flexibility. CircuitSynth~\cite{vijayaraghavan2024circuitsynth} and similar works~\cite{icml2024} target small circuits (up to six components). AnalogCoder~\cite{lai2024analogcoder} generates PySpice code via prompt engineering but depends on a fixed synthesis library and lacks iterative refinement, limiting exploration of novel or complex topologies, particularly for power converter circuits requiring high efficiency and specific output voltage constraints. This limitation makes it less extensible for optimizing circuit performance or handling diverse design constraints. Artisan~\cite{chen2024artisan} is another recent effort focusing on operational amplifier design using domain-specific LLMs. While valuable, it is highly specialized and does not generalize to other circuit families such as power converters. LaMAGIC~\cite{icml2024} fine-tunes LLMs for netlist generation but omits iterative or performance-driven optimization, restricting adaptability to multi-objective constraints. Auto-SPICE~\cite{bhandari2024auto} automates large-scale SPICE netlist generation from textbook schematics (e.g., Masala-CHAI) but focuses on data creation rather than optimization or refinement. Recent advances like AnalogXpert~\cite{zhang2024analogxpert} and Atelier~\cite{shen2024atelier} incorporate domain knowledge, subcircuit libraries, Chain-of-Thought prompting, and agent-based coordination. Nonetheless, these methods lack reinforcement learning for iterative, performance-driven refinement, focusing instead on structured generation and error correction. Our work advances beyond prior efforts by synthesizing more complex circuits while optimizing both topology and performance metrics.

In this work, we present \textsc{AutoCircuit-RL} (\textsc{Ac-Rl}), a reinforcement learning (RL)-based framework that refines LLM-generated circuit topologies to optimize design objectives. Our method employs two training phases: instruction tuning to generate diverse topologies from prompts, followed by RL-refinement using AI-based reward models that estimate validity, efficiency, and output voltage. This enables scaling, generalization, and multi-objective optimization with minimal manual effort. RL-refinement occurs only during training, not at inference. Empirical results show a $\sim$12\% improvement in validity and $\sim$14\% gain in efficiency over the best-performing LLM baselines, with few-shot generalization beyond 6 components and support for circuits with up to 10 components. Key contributions include: \\
\noindent \textbf{LLM with RL Refinement:} We propose \textsc{AutoCircuit-RL}, a novel RL framework for analog circuit synthesis that targets constraint-driven design.\\
\noindent \textbf{Superior Performance Evaluation:} Our framework's evaluations on 4 and 5-component analog circuits demonstrate superior performance in generating circuits that meet design constraints more effectively than other baseline approaches.\\
\noindent \textbf{Scalability and Generalizability:} Using few-shot fine-tuning, our framework generalizes to 6–10 components even with limited data, highlighting its scalability and adaptability in practical design scenarios.

\section{Problem Statement and Dataset}
\label{sec:datasets}
 Given an input instruction, our goal is to produce a netlist with components and their connections. Each entry in the netlist corresponds to a node in an undirected graph $\mathcal{G}$, with edges indicating the connections between these nodes as in the netlist. For the choices of encoding the netlist textually, we adopt the ``Incident" encoding strategy, recognized for its effectiveness in various graph-related tasks \cite{fatemi2024talk}. The center part of Figure \ref{fig:overview} illustrates an example of how a netlist is encoded. This research investigates different model variations that refine Language Models (LMs) to generate the circuit topology netlist. We compare two representation approaches: generating the netlist as lists or employing the text representation with the incident encoding method. Through empirical evaluation, our objective is to evaluate the efficacy of these LM variations in accurately and efficiently synthesizing circuit topologies from natural language instructions.

\begin{figure*}
\centering
\includegraphics[scale=0.668]{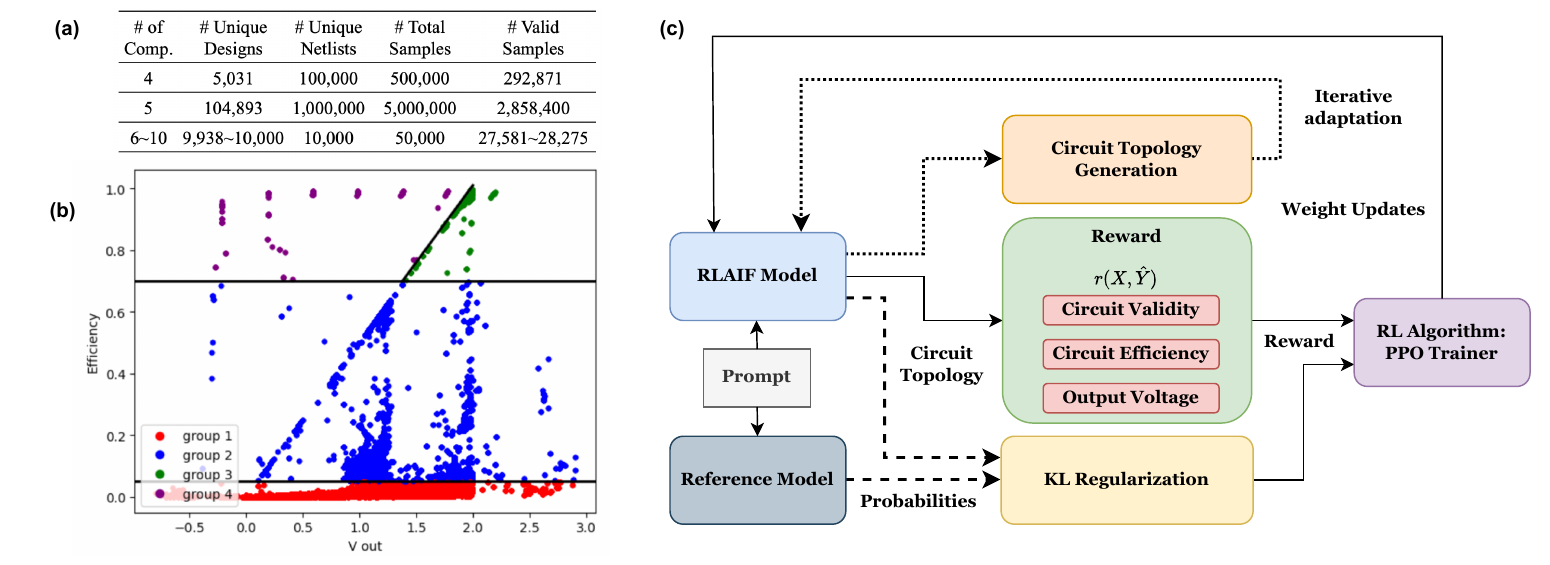}
\caption{((a) Statistics of our Circuit dataset. b) Vout-Efficiency graph of 4-component circuits in the training data. (c) Illustration of the \textsc{AutoCircuit-Rl} framework. Dashed lines depict probability flow, while the dotted line represents iterative adaptation post RLAIF tuning. A prompt guides topology generation, then rewards are calculated with KL-divergence to a reference model to preserve the original distribution. PPO updates model parameters using rewards, and iterative adaptation further improves the model.}
\label{fig:circuitsynth_rl}
\end{figure*}

We generated a dataset of switching power converter topologies with 4–10 components (Figure \ref{fig:circuitsynth_rl}(a)). Using Random Search (RS) \cite{ICCAD21}, we generated numerous unique netlists. Multiple netlists can represent the same topology by changing component order or node indices, so we report unique designs. The design space for 4- and 5-component circuits is small, allowing near-exhaustive exploration; for 6+ components, exhaustive search is impractical. Therefore, we collected 10,000 unique netlists for each component count from 6 to 10. Each netlist was simulated at 5 duty cycles: 0.1, 0.3, 0.5, 0.7, and 0.9, yielding 5 times the number of unique netlists as total samples (Figure~\ref{fig:circuitsynth_rl}(a)). We used NGSpice \cite{nenzi2011ngspice} to identify valid circuits and collect output voltage and efficiency values. The circuit includes five external signal ports: Vin, Vout, GND, N-type gate signal, and P-type gate signal (renamed as `IN', `OUT', `0', `GATEN', and `GATEP', respectively). Devices considered are capacitors, inductors, n-type MOSFET (FET-A), and p-type MOSFET (FET-B). Capacitors and inductors have two ports, while MOSFETs have four (drain, gate, source, body). To simplify the design space and accelerate RS generation, FET-A connects gate/body to \texttt{GATEN}/\texttt{0}, and FET-B to \texttt{GATEP}/\texttt{IN}. Devices are numbered, with shared ports using one index (Figure~\ref{fig:overview}). Capacitors (10$\mu$F), inductors (10$\mu$H), and MOSFETs use fixed parameters; switching frequency is 200\,kHz, input voltage 2V.

For training, we randomly sample approximately 100,000 unique netlists (for 4- and 5-component circuits), each with varying efficiency and output voltage values. The data is categorized into four groups: (a) Group 1 with low efficiency (efficiency $<$ 0.05), (b) Group 2 with moderate efficiency (efficiency between 0.05 and 0.7), (c) Group 3 with high efficiency but minimal output voltage difference from input, and (d) Group 4 with optimal Vout and efficiency. An example for 4-component circuits is shown in Figure \ref{fig:circuitsynth_rl}(b). We apply a weighted sampling strategy in each batch, prioritizing Group 4 (weight 0.4) and giving the lowest priority to Group 1 (weight 0.1). This ensures that batches are enriched with data from Group 4, improving model learning on the most desirable conditions. Instruction prompts are constructed in three categories based on design constraints: (a) Component constraint (only the list of components), (b) Efficiency constraint (components with expected efficiency), and (c) Output Voltage constraint (components with input voltage and expected output voltage).




\section{Proposed Approach}
\label{sec:models}


Our proposed framework, \textsc{AutoCircuit-Rl}, is depicted in Figure \ref{fig:circuitsynth_rl}(c). 
It consists of two primary phases: instruction tuning and RL refinement. In the instruction tuning phase, we fine-tune a large language model (LLM) using supervised learning techniques. This phase focuses on training the model to comprehend instruction prompts that specify the component pool and design constraints, facilitating the efficient generation of valid circuit topologies. To further optimize the circuit topology generation process and ensure compliance with all design constraints, we incorporate reinforcement learning with AI feedback (RLAIF) \cite{bai2022constitutional,lee2023rlaif}. The RL refinement phase enhances circuit topology generation by integrating feedback from constraint-specific AI models in three steps: reward modeling, RL training, and iterative adaptation.


\subsection{Instruction Tuning}
\label{sec:sft}
The instruction tuning phase can be considered as a standard supervised finetuning (SFT) step, where instruction prompts specifying the component pool and design constraints are provided as input to the model, and the corresponding circuit topology generations are produced as output. We represent the pairs of instruction-circuit topologies as $\mathcal{D} = \{(X_i, Y_i)\}_{i=1}^{T}$, where $X_i$ denotes the instruction prompt and $Y_i$ represents the valid circuit topology netlist using the incident encoding method concatenated with the corresponding duty cycle. This phase involves training an autoregressive language model $p_{\theta}$ parameterized by $\theta$ to minimize the negative log-likelihood of the desired circuit topology represented using the incident encoding method. Formally,

\begin{equation}
\mathcal{L}_\mathrm{SFT} = -\mathbb{E}_{(X,Y)\sim{\mathcal{D}}}\left[\sum_r{\log\, \pi_{\theta}(y_t|X,y_{{<t}})}\right]
\end{equation}

where $\pi_\theta$ represents the LLM policy and $y_{<t}$ denotes all tokens before the $t^{th}$ token in the circuit topology $Y$. This objective aims to ensure that the model learns to comprehend the instruction and generate the circuit topology in the incident encoding method. However, the generated circuit topology may not satisfy all the design constraints related to components, efficiency, validity, and expected output voltage. To meet such constraints, we utilize Reinforcement Learning with AI Feedback (RLAIF), which learns to refine the circuit topology generation process by maximizing rewards associated with specific constraints of interest.

\subsection{RL-Refinement with AI Feedback}
The RL-refinement phase aims to enhance the circuit topology generation process by leveraging feedback from constraint-specific AI models. It consists of 3 main steps: 

\subsubsection{Reward Modeling}
In this step, our reward model evaluates the appropriateness of a generated circuit topology based on the instruction prompt. Canonical reinforcement learning with AI feedback (RLAIF) trains a reward model using labeled preferences in the form of triples $\left<X, Y_p, Y_r\right>$, representing the input prompt, preferred topology, and non-preferred topology, respectively. Recent research suggests that directly using reward scores yields better performance than the traditional RLAIF approach. To achieve this, we employ a reward function that evaluates how well the generated circuit topology adheres to different design constraints, such as circuit validity, efficiency, and expected output voltage. To implement this, we train different estimators ($f_\mathrm{clf}$ for all $\mathrm{clf} \in \{\mathrm{valid, eff, vout}\}$) to assess circuit validity, efficiency, and expected output voltage, and assign corresponding scores $s_\mathrm{clf}$. These scores are then used to compute a reward in the range $[-1, 1]$. The estimators are built using dedicated classification or regression models ($f_\mathrm{clf}$), with RoBERTa as the underlying model architecture. Each model is trained on synthetic datasets tailored specifically for each constraint type. The details are as follows:\\
\noindent \textbf{Circuit Validity Estimator} 
A binary classifier $f_\mathrm{valid}$ is trained on a dataset $\mathcal{D}$ comprising valid and invalid circuits, constructed from the aggregated dataset (explained in the Section \ref{sec:datasets}), to determine the validity of circuit topologies. This RoBERTa-based classifier achieves a $92\%$ $F_1$ score for binary classification of circuit validity.

\noindent \textbf{Circuit Efficiency Estimator:} 
A regression model $f_\mathrm{eff}$ estimates circuit efficiency using a subset of the dataset $\mathcal{D}$ with prompts specifying efficiency requirements. The model achieves an $83\%$ macro $F_1$ score by categorizing predicted efficiency scores into predefined categories.

\noindent \textbf{Output Voltage Estimator:} 
A regression model $f_\mathrm{vout}$ predicts output voltage based on input parameters, achieving a low MSE loss of $8e^{-3}$ on the development set.


Using the remaining data in $\mathcal{D}$ and the trained estimators, we define a reward function $r(X, \hat{Y})$ that assigns a reward to an LLM-generated circuit topology $\hat{Y}$ as follows:

\begin{equation}
    r(X, \hat{Y})=
    \begin{cases}
      -1, & \text{if}\ s_\mathrm{valid} < 0.6 \\
      1 , & \text{if}\ s_\mathrm{eff} \text{ or } s_\mathrm{vout} \text{~meets constraints} \\
      s_\mathrm{eff}, & \text{otherwise}
    \end{cases}
\end{equation}

Invalid topologies receive a negative reward. Valid topologies meeting output voltage or efficiency constraints get a reward of 1; otherwise, the efficiency estimate is used as the reward to maximize efficiency.

\subsubsection{RL Tuning}
\label{sec:rl_tuning}
To enhance the LLM for generating circuit topologies that better meet the constraints, we employ a reward function $r(X,\hat{Y})$ and Proximal Policy Optimization (PPO) \cite{schulman2017proximal}. The base model for this refinement is the LLM fine-tuned with the instruction tuning technique discussed in Section \ref{sec:sft}, following the common practice in Reinforcement Learning with Human Feedback (RLHF) \cite{ouyang2022training}. Standard PPO training procedures are then applied to optimize the base model using the following reward objective function:
\begin{gather}
\mathcal{L}_{RL} = r(X, \hat{Y}) - \eta KL(\pi_\mathrm{RLAIF}(\hat{Y}|X) || \pi_{\theta}(\hat{Y}|X))
\label{eq:rl}
\end{gather}
Here, KL represents the Kullback-Leibler divergence, and $\eta$ is a hyperparameter controlling the penalty for divergence. This penalty helps prevent the model from getting trapped in local optima or straying too far from the original distribution of the supervised instruction-tuned model.

 \subsubsection{Iterative Adaptation (IA)}
 \label{sec:ia}
Additionally, we explore the concept of iterative adaptation as a means to refine the circuit topology generation process further. Leveraging the circuits generated in the previous steps, we aim to enhance the overall synthesis task by iteratively adapting the model based on the sampled valid and highly efficient synthesized circuit topologies. Utilizing the nucleus sampling approach, we specifically target circuit topologies with an efficiency score ($s_\mathrm{eff}$) exceeding 0.7. This criterion ensures that the selected circuits not only satisfy validity constraints but also exhibit high operational efficiency. To initiate the iterative adaptation process, we synthesize a dataset comprising 10,000 such samples. Starting from the RL-tuned model described in Section~\ref{sec:rl_tuning}, we iteratively refine the model using Equation~\ref{eq:rl}, incorporating insights from high-quality topologies into the RL-refined model from the previous iteration. Upon completion, the final model is used directly for inference, with no further tuning. In our experiments, we evaluate this iterative adaptation's effectiveness to enhance the circuit topology generation process. 


\section{Experiments and Results}

\begingroup
\renewcommand{\arraystretch}{1.3} 
\setlength{\tabcolsep}{4pt} 
\begin{table*}[]
\footnotesize

\centering
\caption{Evaluation results for all methods on 4-component (4C) and 5-component (5C) circuits. Circuit validity and efficiency are measured using both classifier (column 2 and 4) and simulator (column 3 and 5). The DGR $\rho$ and success rate $\sigma$ for different categories of constraints: component (C), efficiency (C+E), output voltage (C+V) and overall (O) are also reported.}
\begin{tabular}{lcccccccccccccc}
\hline
\multicolumn{1}{l|}{Models}                      
& \multicolumn{2}{c|}{$E(f_\mathrm{valid}(\hat{y}))$}       
& \multicolumn{2}{c|}{$E(f_{\mathcal{S}_\mathrm{valid}}(\hat{y}))$} 
& \multicolumn{2}{c|}{$E(f_\mathrm{eff}(\hat{y}))$}         
& \multicolumn{2}{c|}{$E(f_{\mathcal{S}_\mathrm{eff}}(\hat{y}))$} 
& \multicolumn{2}{c|}{DGR $\rho$} 
& \multicolumn{4}{c}{Success Rate $\sigma$ (\%)} 
\\ \hline
\multicolumn{15}{c}{Prompt Tuning}                            \\ \hline
\multicolumn{1}{l|}{}                                     & 4C      & \multicolumn{1}{c|}{5C} & 4C          & \multicolumn{1}{c|}{5C}     & 4C      & \multicolumn{1}{c|}{5C} & 4C          & \multicolumn{1}{c|}{5C}    & \multicolumn{2}{c|}{}       & C           & C+E         & C+V     & O          \\ \hline
\multicolumn{1}{l|}{Llama-2-13b$_\mathrm{p_{100}}$}         & 54.60                & \multicolumn{1}{c|}{50.20}           &      54.50                    & \multicolumn{1}{c|}{53.40}                    & 53.60                 & \multicolumn{1}{c|}{43.62}            &          52.45                & \multicolumn{1}{c|}{47.97}                   & \multicolumn{2}{c|}{5.39}                  &           80.94                   &    33.89                                   &            34.02    & 43.35            \\
\multicolumn{1}{l|}{Flan-UL2-20b$_\mathrm{p_{100}}$}       & \textbf{57.40}                & \multicolumn{1}{c|}{\textbf{51.60}}           &     \textbf{57.35}                    & \multicolumn{1}{c|}{\textbf{56.00}}                    & \textbf{56.20}                 & \multicolumn{1}{c|}{\textbf{44.50}}            &      \textbf{56.08}                    & \multicolumn{1}{c|}{\textbf{49.64}}                   & \multicolumn{2}{c|}{\textbf{4.76}}                  &     \textbf{84.09}                         &     \textbf{36.54}                                  &            \textbf{34.67}              & \textbf{45.30}    \\ \hline
\multicolumn{14}{c}{Vanilla \& Gumbel-based Multi-Objective Fine Tuning}            \\ \hline
\multicolumn{1}{l|}{GPT-Neo$_\mathrm{FT}$}                       & 60.50                & \multicolumn{1}{c|}{57.80}           &    60.60                      & \multicolumn{1}{c|}{57.78}                    & 57.90                 & \multicolumn{1}{c|}{54.70}            &      56.73                    & \multicolumn{1}{c|}{54.49}                   & \multicolumn{2}{c|}{2.98}                  &      89.52                        & \multicolumn{1}{c}{40.28}                  &       37.62                 & 49.06      \\
\multicolumn{1}{l|}{StableLM$_\mathrm{FT}$}                      & 59.42                & \multicolumn{1}{c|}{58.10}           &           60.20               & \multicolumn{1}{c|}{59.00}                    & 58.20                 & \multicolumn{1}{c|}{55.0}            &            59.51             & \multicolumn{1}{c|}{53.27}                   & \multicolumn{2}{c|}{2.67}                  &      89.04                        & \multicolumn{1}{c}{39.76}                  &        37.96  & 48.89                    \\
\multicolumn{1}{l|}{Llama-3$_\mathrm{FT}$}                         & {66.78}                & \multicolumn{1}{c|}{{63.50}}           &        { 66.80}                 & \multicolumn{1}{c|}{{63.80}}                    & {62.8 }                & \multicolumn{1}{c|}{{61.3}}            & \multicolumn{1}{c}{{61.15}}     & \multicolumn{1}{c|}{{62.97}}                  & \multicolumn{2}{c|}{\textbf{2.10}}                  & \multicolumn{1}{c}{{95.76}}         &    {68.04}                                    & \multicolumn{1}{c}{58.06} & {69.79}        \\
\multicolumn{1}{l|}{MPT-7b$_\mathrm{FT}$}                          & 64.96                & \multicolumn{1}{c|}{61.65}           &              65.00            & \multicolumn{1}{c|}{60.50}                    & 60.50                 & \multicolumn{1}{c|}{59.2}            & \multicolumn{1}{c}{60.23}     & \multicolumn{1}{c|}{62.23}                   & \multicolumn{2}{c|}{2.26}                  & \multicolumn{1}{c}{95.32}         &   67.85                                    & \multicolumn{1}{c}{{58.28}}  & 69.52  \\
\multicolumn{1}{l|}{Gumbel$_\mathrm{Llama-3}$}                          & \textbf{67.60}           & \multicolumn{1}{c|}{{\textbf{65.75}}}           &              \textbf{67.32 }           & \multicolumn{1}{c|}{{\textbf{64.19}}}                    & \textbf{67.15 }                & \multicolumn{1}{c|}{64.38}            & \multicolumn{1}{c}{{\textbf{63.87}}}     & \multicolumn{1}{c|}{63.15}                   & \multicolumn{2}{c|}{2.19}                  & \multicolumn{1}{c}{\textbf{96.04}}         &   \textbf{69.56}                                    & \multicolumn{1}{c}{\textbf{60.36}}  & \textbf{71.27}  \\
\multicolumn{1}{l|}{Gumbel$_\mathrm{MPT-7b}$}                          & 67.16                & \multicolumn{1}{c|}{63.50}           &              66.42            & \multicolumn{1}{c|}{63.48}                    & 66.30                 & \multicolumn{1}{c|}{\textbf{64.42}}            & \multicolumn{1}{c}{63.19}     & \multicolumn{1}{c|}{\textbf{63.64}}                   & \multicolumn{2}{c|}{2.32}                  & \multicolumn{1}{c}{95.80}         &   68.30                                    & \multicolumn{1}{c}{{60.22}}  & 70.68
\\ \hline
\multicolumn{14}{c}{\textsc{AutoCircuit-Rl} (our approach)}                          \\ \hline
\multicolumn{1}{l|}{\textsc{Ac-Rl}$_\mathrm{Llama-3}$} & \textbf{\underline{75.11}}       & \multicolumn{1}{c|}{\textbf{\underline{{73.46}}}}  &    \textbf{\underline{74.48}}                      & \multicolumn{1}{c|}{\textbf{\underline{73.96}}}           & \textbf{\underline{74.20}}        & \multicolumn{1}{c|}{\textbf{\underline{73.60}}}   & 71.65                & \multicolumn{1}{c|}{\textbf{\underline{72.22}}}          & \multicolumn{2}{c|}{\textbf{\underline{1.29}}}         & \textbf{\underline{99.08}}                    & \multicolumn{1}{c}{\textbf{\underline{80.90}}}         &             71.30     & \textbf{\underline{80.69}}           \\
\multicolumn{1}{l|}{\quad w/o IA}           & 71.08                & \multicolumn{1}{c|}{68.32}           &      72.78                       & \multicolumn{1}{c|}{68.96}                    & 71.60                 & \multicolumn{1}{c|}{69.70}            &             69.50             & \multicolumn{1}{c|}{68.68}                   & \multicolumn{2}{c|}{1.46}                  &       98.26                     & \multicolumn{1}{c}{76.75}                  &          70.10      & 78.39              \\
\multicolumn{1}{l|}{\textsc{Ac-Rl}$_\mathrm{MPT-7b}$}  & 74.20                & \multicolumn{1}{c|}{72.65}           & 74.32                & \multicolumn{1}{c|}{72.34}                    & 73.40                 & \multicolumn{1}{c|}{72.70}            &        \textbf{\underline{72.08 }}                 & \multicolumn{1}{c|}{71.94}                   & \multicolumn{2}{c|}{1.34}                  &               99.05               & \multicolumn{1}{c}{79.85}                  & \textbf{\underline{71.64}}    & 80.41                 \\
\multicolumn{1}{l|}{\quad w/o IA}           & 70.88                & \multicolumn{1}{c|}{69.56}           &       71.06                   & \multicolumn{1}{c|}{69.23}                    & 70.10                 & \multicolumn{1}{c|}{69.50}            &          67.37                & \multicolumn{1}{c|}{69.08}                   & \multicolumn{2}{c|}{1.53}                  &       98.03                       & \multicolumn{1}{c}{76.52}                 &        68.50    & 77.61                  \\ \hline
\end{tabular}

\label{tab:eval_res}

\end{table*}
\endgroup

In this study, we utilized the following Language Models (LMs) for generating circuit topologies: GPT-Neo-2.7 \cite{black2021gpt}, StableLM-3B-4E1T, Llama-3-8b \cite{dubey2024llama}, and MPT-7b \cite{mosaicml2023introducing}. More details on the baseline models and implementation are provided in the Appendix \ref{sec:baseline}, and \ref{app:implement}.

\subsection{Baselines}

Our primary focus is on leveraging LLM-based generative methods for circuit design synthesis and benchmarking their potential. We compare different LLM-based methods as baselines and include GraphVAE~\cite{simonovsky2018graphvae}, a non-LLM baseline, to highlight the superiority of our method in both efficiency and performance. While recent LLM-based frameworks like AnalogCoder~\cite{lai2024analogcoder} and Artisan~\cite{chen2024artisan} provide valuable contributions, their methodologies and application domains differ from ours. AnalogCoder employs a training-free prompt strategy with a fixed synthesis library, focusing on general analog circuits, and does not specifically address power converters, which require nuanced handling of constraints such as efficiency and output voltage. Artisan is tailored for operational amplifier design using a domain-specific LLM. In contrast, \textsc{AutoCircuit-RL} is currently trained using power converter designs and combines instruction tuning with RL refinement to handle diverse user prompts and optimization goals. Although these prior works are not directly applicable to our constraint-driven setting, future adaptations could make comparisons more feasible.

Given these considerations, we evaluate \textsc{AutoCircuit-RL} against the following baselines:
\textbf{Zero-Shot Generation}: Prompts with component pools and design constraints are directly fed into large language models (LLMs) like Llama-2 (13b) and Flan-UL2 (20b) without fine-tuning, aiming to generate circuit topologies;  
\textbf{In-Context Learning (ICL)}: This approach uses circuit generation demonstrations, combining input prompts with component pools, design constraints, and corresponding output circuits within the prompts. It leverages the in-context learning ability of LLMs such as Llama-2 (13b) and Flan-UL2 (20b), exploring different numbers of examples ($j \in \{5, 10, 20\}$) and experimenting with incident encoding and netlist structures;  
\textbf{Prompt Tuning}: This method fine-tunes LLMs like Llama-2 (13b) and Flan-UL2 (20b) for circuit topology generation using a Prompt-tuned Model \cite{lester2021power}, which learns task-specific soft prompts while keeping model parameters unchanged. We test with 100 trainable soft prompt tokens (p100);  
\textbf{Vanilla Fine-Tuning}: Standard fine-tuning is conducted on the LMs listed above. The primary objective is to minimize the negative log-likelihood for generating circuit topologies;  
\textbf{Gumbel-Max Fine-Tuning}: Drawing ideas from a prior study \cite{vijayaraghavan2024circuitsynth}, we integrate multiple objectives optimizing for circuit validity and efficiency using the Gumbel-Max trick. This approach refines models fine-tuned using Llama-3 (8b) and MPT-7b architectures, allowing for more efficient circuit generation while maintaining structural validity constraints;  
\textbf{\textsc{GraphVAE}}: This method models circuit netlists as undirected graphs and uses a variational autoencoder (VAE) to generate graphs from continuous embeddings \cite{simonovsky2018graphvae}. We did not implement DAG-based methods\cite{dong2023cktgnn, zhang2019d}, because a noticeable subset of circuits in our datasets are not DAGs. For circuits with 6–10 components, the generation process is conditioned on a SentenceBERT-encoded label vector \cite{reimers2019sentence} derived from the input prompt, enabling controlled and guided sampling during inference;  
\textbf{\textsc{AutoCircuit-RL}}: We introduce \textsc{AutoCircuit-RL}, a comprehensive framework designed to enhance the circuit topology generation process using RL and iterative adaptation.

\subsection{Metrics}
In our evaluation setup, we report different metrics based on sampling 500 unique circuit topologies from each of the trained models.  \textbf{(a) Circuit Validity Score} represents the fraction of unique circuit topologies estimated as valid, denoted as $E(f_\mathrm{clf}(\hat{y}))$. Here, $\mathrm{clf} \in \{\mathrm{valid}, \mathcal{S}_\mathrm{valid}\}$ refers to the validity estimated by the classifier or the simulator, respectively. We consider a circuit valid if its validity score from the classifier exceeds 0.6; \textbf{(b) Circuit Efficiency Score} refers to average efficiency of the generated circuits using our efficiency regressor or the NGSpice simulator, denoted as $E(f_\mathrm{eff}(\hat{y}))$ and $E(f_{\mathcal{S}_\mathrm{eff}}(\hat{y}))$, respectively. The NGSpice simulator evaluates the validity and efficiency of a given netlist by verifying its electrical characteristics and performance metrics through detailed circuit simulations. It evaluates parameters such as voltage levels, timing, and power consumption of the circuit topology with certain duty cycles under given conditions; \textbf{(c) Duplicate Generation Rate (DGR)}, denoted by $\rho$, indicates the number of circuit topologies required to be sampled from the model to obtain a unique circuit topology design. Formally, $\rho$ is calculated as the number of topologies generated divided by the number of unique topologies (500 in our case); and \textbf{(d) Success Rate} computes the percentage of valid circuit topologies that successfully meet certain design constraints for each category of prompts (as in Section \ref{sec:datasets}): Component constraint (C), Efficiency constraint (C+E), Output Voltage constraint (C+V) and overall success rate (O). Figure \ref{fig:overview} presents a sample success scenario with efficiency constraint.
\subsection{RL Convergence Analysis}
\label{sec:rl_convergence}

\begin{figure}[!ht]
\centering
\includegraphics[width=0.98\linewidth]{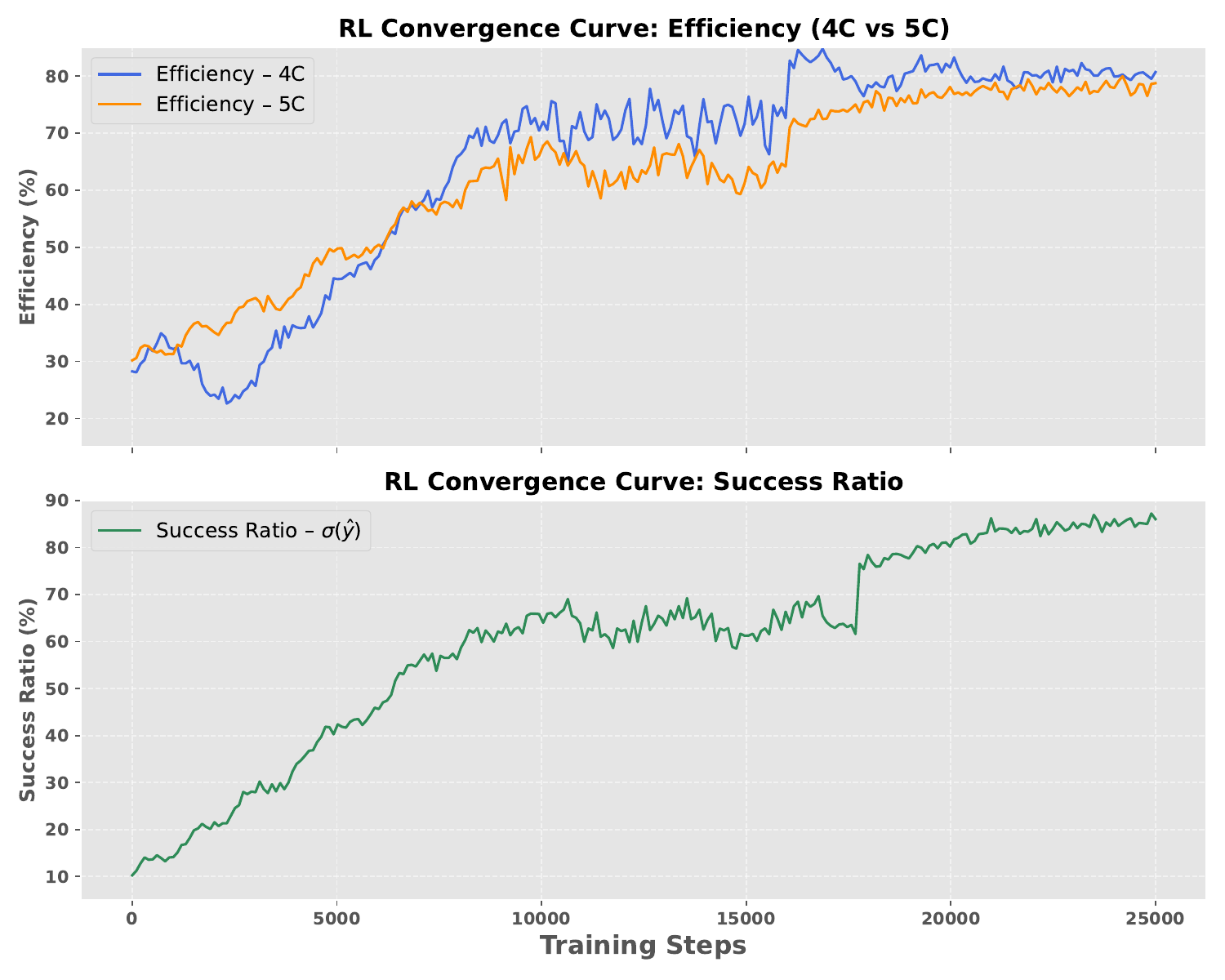}
\caption{RL convergence curves for \textsc{AutoCircuit-RL} with Llama-3 over $\sim$25,000 training steps. Top: Efficiency for 4- and 5-component circuits; Bottom: Success ratio of valid and constraint-satisfying topologies.}
\label{fig:rl_convergence}
\end{figure}

We evaluate the learning dynamics of the proposed \textsc{AutoCircuit-RL} framework by analyzing the convergence behavior of two key metrics during RL refinement: (a) circuit efficiency and (b) success ratio, defined as the proportion of generated circuits satisfying functional and design constraints. Figure~\ref{fig:rl_convergence} illustrates these convergence curves for circuits composed of 4 and 5 components, trained using the Llama-3 backbone over $\sim{25,000}$ training steps with a batch size of 16. The training exhibits several characteristic phases. In the initial phase, both efficiency and success ratio increase gradually with noticeable oscillations, reflecting the model's early adaptation to reward signals. This is followed by a phase of rapid improvement, where the RL agent learns effective circuit design strategies. Intermediate fluctuations arise as the model explores diverse topologies while balancing validity, constraint satisfaction, and efficiency. Eventually, the curves stabilize and plateau, indicating convergence to a policy that consistently generates valid, high-quality circuit topologies. These results show \textsc{AutoCircuit-RL} steadily improves design capabilities and sustains robust performance as circuit complexity grows, with only minor efficiency degradation observed from 4- to 5-component circuits.


\subsection{Results Overview}
The evaluation results, summarized in Table~\ref{tab:eval_res}, show that \textsc{AutoCircuit-RL} outperforms all baselines across various metrics for circuit topology synthesis. Notably, smaller language models tuned with our method outperform larger prompt tuning-based models, generating unique designs faster with a higher success rate in meeting design constraints. Our LLM-based approach is efficient, requiring $\sim{2.7}$ seconds per design generation once trained (refer Section \ref{app:runtime} for runtime analysis). In contrast, traditional search-based algorithms \cite{ICCAD21} are computationally expensive and time-consuming, often taking hundreds of seconds to converge on a target design.\\
\textbf{Usability of Zero-Shot and ICL Methods}
We observe a notable disparity in performance between zero-shot generation techniques and fine-tuned methodologies. Zero-shot generation struggles to produce valid netlist-like structures essential for subsequent classification or simulation tasks. Despite using in-context learning (ICL), where the model is exposed to sample prompts and corresponding circuit generations, the improvement in generating comprehensive netlist-like structures is minimal. Additionally, increasing the number of in-context examples $j$ yielded diminishing returns. As a result, these incomplete structures cannot be used for accurate metric computation, so we exclude the zero-shot and ICL results in Table~\ref{tab:eval_res}.

\subsection{Error Analysis}
We conduct a qualitative error analysis of the \textsc{AutoCircuit-RL} framework by evaluating the model-generated circuits through post-hoc SPICE simulations. The analysis focuses on two primary failure modes: validity errors, where circuits fail to simulate due to structural violations such as incorrect node assignments or connectivity issues, and efficiency constraint errors, where the generated circuits do not meet the efficiency thresholds specified in the generation prompts. This systematic evaluation enables the identification of recurring failure patterns and provides insight into the model's behavior near constraint boundaries, highlighting specific areas where the model lacks fine-grained control.

Our analysis reveals that validity failures predominantly arise from minor inconsistencies in node assignments rather than fundamental topological errors, and such failures can typically be resolved with minimal structural modifications. Regarding efficiency constraint errors, most of the circuits that fail to meet the specified thresholds do so by a relatively small margin, suggesting that the model generally approximates the desired performance metrics. These deviations often reflect inherent trade-offs with other design parameters such as output voltage and duty cycle. A limited number of outlier cases exhibit larger deviations from the efficiency targets, typically caused by complex interactions between circuit components. Collectively, these findings indicate that while the model internalizes key principles of both structural integrity and performance-aware design, there remains scope for further improvements in constraint calibration and optimization. Detailed examples and further discussion can be found in Appendix~\ref{sec:error-analysis}.

\subsection{Effectiveness of \textsc{AutoCircuit-Rl}}

\textbf{Impact of Prompt Tuning}
Experiments with prompt tuning of Flan-UL2/Llama-2 models (approximately 20b parameters) with a restricted dataset, facilitated the production of netlist-like structures. This represents a substantial enhancement compared to the performance of the same models under zero-shot or in-context learning (ICL) conditions. Nevertheless, these fine-tuned models lag behind smaller language models (such as GPT-Neo/StableLM) fine-tuned using our method in terms of both efficiency and validity of the generated circuits. Additionally, our models demonstrate a higher success rate in meeting constraints compared to the larger prompt-tuned language models. Despite the advantages of fine-tuning these larger models, we emphasize that fine-tuned models with lower capacity can still achieve effective performance relative to larger prompt tuning-based models for circuit topology synthesis.

\textbf{Comparison with Vanilla Fine-tuning}
When contrasting our methodology, which entails iterative refinement via reinforcement learning, with the basic fine-tuning of different architectures, we note a significant enhancement in the effectiveness of our approach. Our methodology, which entails iterative refinement via reinforcement learning, demonstrates a significant improvement in effectiveness compared to basic fine-tuning of GPT-Neo and StableLM architectures. This improvement is evident in generating valid circuits that meet design constraints, underscoring the efficacy of our approach in synthesizing valid topologies and accelerating the discovery of unique designs.


\textbf{Effect of Using Gumbel-Max Trick for Multi-Objective Optimization}  
Building on a prior study \cite{vijayaraghavan2024circuitsynth}, we optimize circuit validity and efficiency using the Gumbel-Max trick, yielding notable gains over standard fine-tuned models. However, its performance still lags behind \textsc{AutoCircuit-RL} by $\sim9\%$ in circuit validity and efficiency. A key limitation of the Gumbel-Max trick is its inability to adaptively optimize multiple objectives. Unlike reinforcement learning (RL), which refines strategies through iterative AI feedback, Gumbel-based methods don't adjust based on prior evaluations and lack a mechanism to balance competing objectives, often leading to suboptimal designs. In contrast, our RL approach navigates these trade-offs effectively, explores a broader design space, and avoids premature convergence, demonstrating superior performance in circuit validity and efficiency.



\textbf{Effect of Iterative Adaptation}
To assess the significance of the iterative adaptation strategy for our task, we conducted an experiment where we sampled circuit topology generations from a model trained using reward models but without incorporating the iterative adaptation strategy. Our results reveal that the iterative refinement enabled by iterative adaptation substantially improves the overall model performance across various metrics. Notably, the lack of an iterative adaptation strategy leads to a more pronounced decline in performance, particularly notable in the circuit efficiency score, where we observed a $\sim8\%$ decrease.

\begin{figure}
\centering
\includegraphics[scale=0.3]{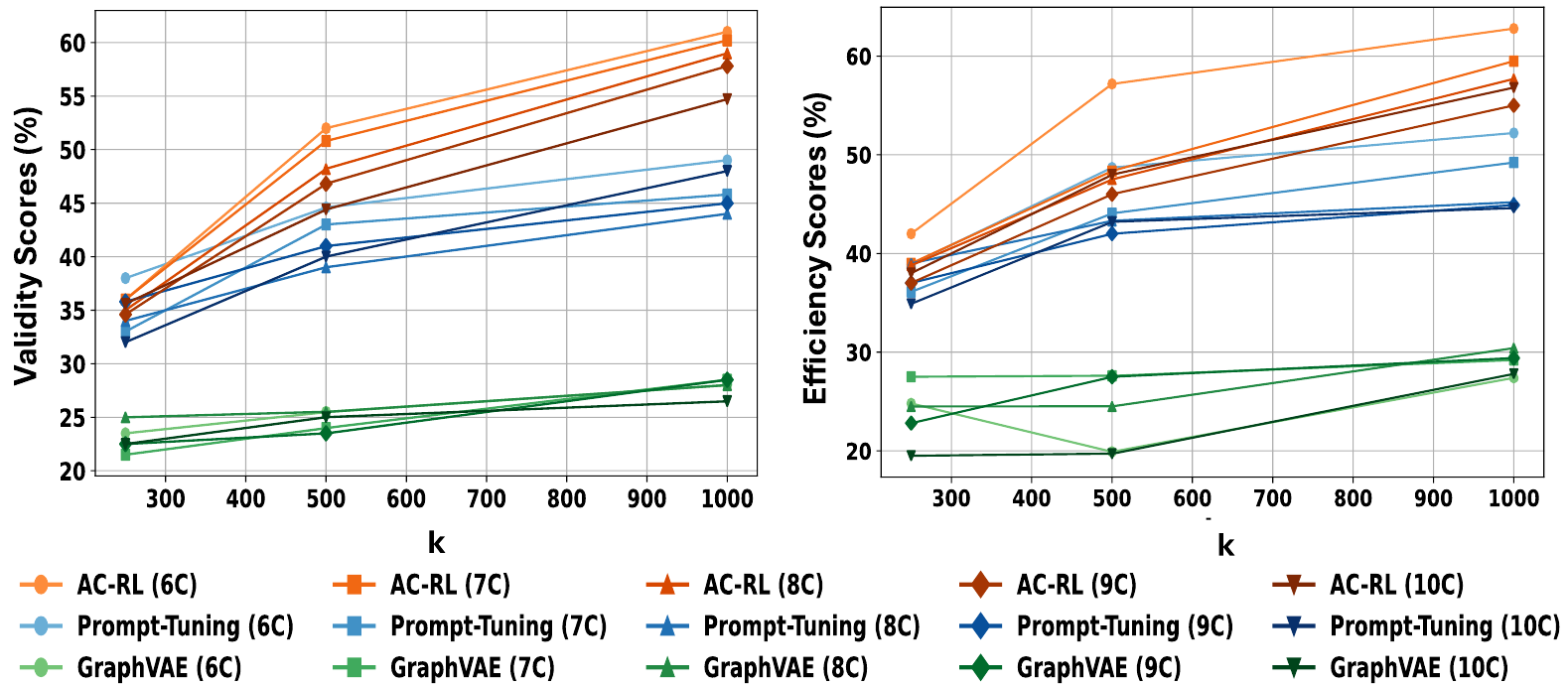}
\caption{Few-shot fine-tuning results for 6-10 components, with models tuned using $k$-examples of varying circuit topologies.}
\label{fig:few_shot}

\end{figure}

\begin{table}[]
\footnotesize
\centering
\caption{Overall success rate of \textsc{GraphVae}, Prompt Tuning and \textsc{AutoCircuit-Rl} for 6-10 component circuits with $k=1000$. }
\label{tab:success_rate_6_10}
\begin{tabular}{@{}lccccc@{}}
\toprule
\multicolumn{1}{c}{Models} & 6C & 7C & 8C & 9C & 10C \\ \midrule
\textsc{GraphVae}                       & 21.5            & 20.6            &    16.1         & 15.9            & 12.8             \\
\textsc{Prompt Tuning}                       & 39.7            & 36.3            & 35.9            & 33.2            & 32.1             \\
\textsc{Ac-Rl}            &\textbf{ 65.5}            & \textbf{63.8}            & \textbf{63.2}            & \textbf{60.4}            & \textbf{58.5 }            \\ \bottomrule
\end{tabular}

\end{table}

\subsection{Adherence to Design Constraints}
This section examines how different models meet three key design constraints for 4- and 5-component circuits. We evaluate success rate $(\sigma)$ for component usage (C), efficiency (C+E), and output voltage (C+V), using a 20-40-40 prompt split per constraint. \textsc{AutoCircuit-Rl} excels across constraints, demonstrating RL's effectiveness in circuit topology synthesis.


\textbf{Component Pool Adherence (C)}
Ensuring adherence to the component pool constraint is pivotal for practical applicability, achievable with minimal tuning. Generally, models fine-tuned or optimized with reward models exhibit superior adherence to the specified component pool. For instance, prompt tuning-based models, even with minimal tuning, achieve moderate success rates, with $\sigma$ values ranging from $\sim{81}\%$ to $\sim{84}\%$. Remarkably, models trained using our complete \textsc{AutoCircuit-Rl} approach showcase the highest success rates, with $\sigma$ values surpassing $\sim{98\%}$.

\begin{figure}[h]
\centering
\includegraphics[scale=0.4]{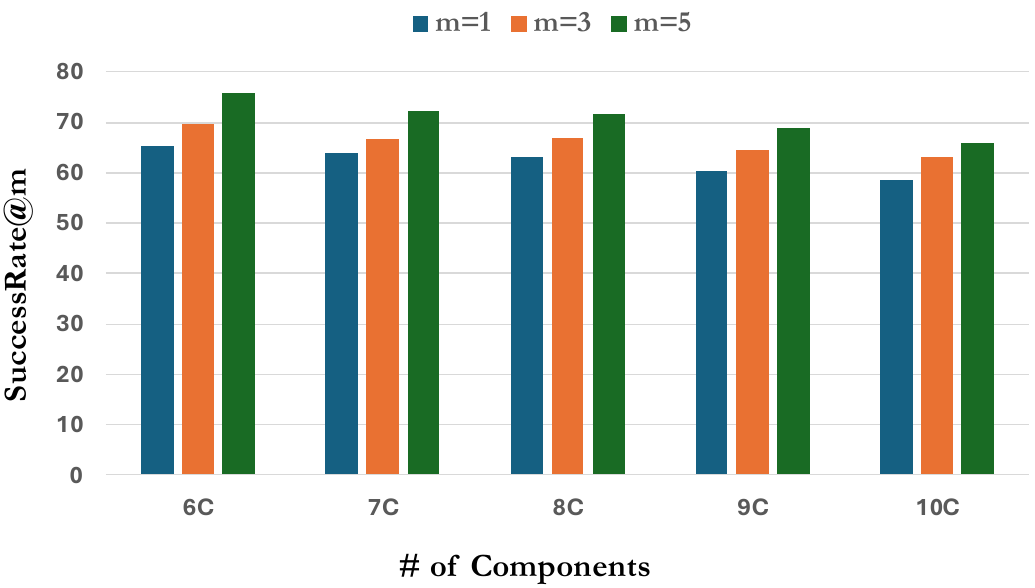}
\caption{Plot of SuccessRate@\(m\) of \textsc{Ac-Rl} for $m\in\{1,3,5\}$.}
\label{fig:srm}
\end{figure} 

\textbf{Efficiency (E) \& Output Voltage (V) Adherence}
Meeting efficiency and expected output voltage constraints is challenging in comparison to the component pool constraint, particularly when expected values are specified in the prompts. However, models optimized with reward models show notable improvements in meeting constraints. Particularly, models trained using \textsc{AutoCircuit-Rl} exhibit superior performance in satisfying efficiency and expected output voltage requirements compared to their counterparts.

\subsection{Generalization to Complex Scenarios}

To assess the generalization capability of our framework, we evaluated it on circuits with 6–10 components. Following the procedure in Section~\ref{sec:datasets}, we constructed a limited dataset and applied few-shot fine-tuning with $k = \{250, 500, 1000\}$ examples. The base model for fine-tuning was the \textsc{Ac-Rl} model previously trained on 4- and 5-component circuits (see Table~\ref{tab:eval_res}). We adopted a sampling strategy similar to that used for 4/5-component circuits, as shown in Figure \ref{fig:circuitsynth_rl}(b). We compared our top-performing \textsc{AutoCircuit-Rl} (Llama-3) approach against \textsc{GraphVAE} and prompt-tuned Flan-ul-20b, focusing on circuit validity and efficiency, measured by the NGSpice simulator. As shown in Figure \ref{fig:few_shot}, \textsc{AutoCircuit-Rl} showed consistent improvements with increasing $k$, achieving validity and efficiency scores nearing 50\% with fewer than 500 examples. Table~\ref{tab:success_rate_6_10} summarizes the success rates for all methods with $k = 1000$, highlighting the superior performance of \textsc{AutoCircuit-Rl}. The higher success rate underscores the effectiveness of reward models in optimizing efficiency and validity. These findings show \textsc{AutoCircuit-Rl}'s potential to scale to higher component circuits using minimal fine-tuning, though addressing more complex design constraints for different circuit types remains a future research area.

\textbf{Analysis of Success Rates \& Sampling Strategies}  
Comparing success rates in Tables \ref{tab:eval_res} and \ref{tab:success_rate_6_10}, we observe a performance decline in \textsc{Ac-Rl} as circuit complexity increases. However, this drop is due to differences in training and evaluation setups, as the model was extensively trained on 4C and 5C circuits, whereas 6C and beyond used a few-shot setup with only 1,000 samples. Despite this, the model achieves \textgreater${60}\%$ success with minimal training data, demonstrating good generalization. As shown in Figure \ref{fig:few_shot}, increasing training data improves performance, reinforcing the model's ability to learn from limited data. We also analyzed \textit{SuccessRate@m}, where multiple generations per prompt (\textit{m}) increase the likelihood of success. As seen in Figure \ref{fig:srm}, SuccessRate@3 and SuccessRate@5 consistently outperform SuccessRate@1, showing that additional sampling improves results. The average generation times for $m=3$ and $m=5$ are $\sim{7}$ and $\sim{9}$ seconds, respectively, achieving higher success rates with minimal added computation and remaining significantly faster than traditional AI-based search methods, which require hundreds of seconds.

\section{Discussion}

Our results demonstrate the effectiveness of proposed \textsc{Ac-Rl} in advancing circuit topology synthesis by iteratively refining circuit generation using reinforcement learning. Its key advantages are as follows:

\textbf{RL for Improved Validity and Efficiency}  
Unlike traditional methods, \textsc{Ac-Rl} adapts circuit generation using reward feedback. This iterative process yields higher validity and efficiency, outperforming zero-shot and ICL baselines. Its ability to optimize multiple objectives while adhering to constraints highlights its robustness.\\
\textbf{Accelerated Discovery of Novel Circuits}
\textsc{Ac-Rl} reduces duplicate generation rates (\textsc{DGR}) by $\sim{38}\%$ compared to fine-tuning methods, enabling faster convergence to unique and effective circuit topologies. This accelerates design space exploration while minimizing redundant computations, making it highly effective for circuit automation.\\
\textbf{Enhanced Constraint Adherence}
With up to 80\% success rates in constraint adherence, \textsc{Ac-Rl} efficiently balances efficiency and output voltage while ensuring feasibility. Unlike heuristic-based methods that require extensive computational resources, its reward-driven approach adapts to complex constraints with significantly better performance.\\
\textbf{Scalability with Limited Training Data}
\textsc{Ac-Rl} achieves over $\textgreater60\%$ success rates generating valid circuits with only $\sim{1,000}$ training examples, demonstrating strong generalization. This ability to perform well with limited data distinguishes it from conventional methods relying on extensive labeled datasets.\\
\textbf{Generalization to Increasing Circuit Complexity}
\textsc{Ac-Rl} scales effectively to circuits with 6-10 components, outperforming baselines like \textsc{GraphVAE} and prompt-tuned \textsc{Flan-UL-20b}. While complexity increases, reducing success rates, more training data significantly enhances performance. Additionally, SuccessRate@m analysis shows that generating multiple circuits per prompt improves convergence with minimal added computation.

Extending \textsc{Ac-Rl} to diverse circuit architectures and integrating RL with advanced sampling could further optimize high-dimensional design spaces. Addressing constraints like power consumption and component parameter estimation via adaptive reward modeling remains a promising avenue. Overall, \textsc{AutoCircuit-RL} offers a transformative approach to circuit topology synthesis, excelling in efficiency, constraint adherence, and generalization with minimal data. Further enhancements could solidify its role as a cornerstone in AI-driven electronic design automation.

\section{Conclusion}
\label{sec:conclusion}

In this work, we introduced \textsc{AutoCircuit-RL} (\textsc{Ac-Rl}), a framework for automating analog circuit topology synthesis via a two-phase training process. The first phase employs instruction tuning, where an LLM generates initial topologies from structured prompts encoding design constraints like component pools and target efficiency or output voltage, ensuring basic feasibility. The second phase, RL refinement, updates the model using AI-based reward feedback to improve validity, efficiency, and output voltage. This refinement is applied only during training to improve generation quality, enabling faster inference. Empirical results show that \textsc{Ac-Rl} outperforms prior methods, generating $\sim12\%$ more valid circuits and improving efficiency by $\sim14\%$ for 4- and 5-component circuits. It demonstrates strong generalization, achieving near 50\% validity and efficiency for larger circuits with minimal training data. Additionally, it reduces duplicate generation by $\sim{38}\%$ compared to other baselines and, with only 1,000 training examples, generates designs where \textgreater60\% meet the specified constraints. \textsc{Ac-Rl}'s reward-driven optimization adapts to evolving design constraints, setting a new benchmark for AI-driven design automation. Our findings aim to accelerate the development of efficient, adaptable methodologies in analog circuit automation, enabling faster, reliable synthesis for circuit topologies with reduced training data needs. Future work will extend \textsc{Ac-Rl} to support more complex designs, incorporate advanced sampling, and refine power models, further advancing electronic design automation.

\section*{Impact Statement}

The proposed \textsc{AutoCircuit-RL} framework significantly advances the field of analog circuit topology synthesis by integrating large language models (LLMs) with reinforcement learning (RL) for constrained topology generation. Our approach addresses key challenges in circuit synthesis, such as scalability, efficiency, and adaptability to diverse design specifications, which are critical in modern electronic design automation (EDA). By overcoming the limitations of traditional search-based and generative methods, our framework enables efficient circuit generation while reducing the number of simulation queries. This not only enhances the practicality of AI-driven circuit design but also provides a scalable solution capable of generalizing to circuits with 6 to 10 components using few-shot fine-tuning. Our results demonstrate superior performance in generating circuits that meet design constraints more effectively than existing approaches, paving the way for future research in topology-aware AI design methods.

Beyond technical improvements, our work has broader societal and ethical implications. By automating complex aspects of circuit design, \textsc{AutoCircuit-RL} reduces reliance on extensive human expertise, potentially lowering the barrier to entry for new designers and expanding accessibility to hardware design in low-resource environments. This democratization of circuit synthesis could foster innovation and accelerate technological advancements across industries. However, it is crucial to ensure that AI-driven design tools remain transparent and do not introduce unintended biases in circuit topology selection, which could lead to over-reliance on specific architectures. Additionally, given the increasing energy demands of AI-driven automation, future work should consider optimizing the computational efficiency of training and inference phases to minimize environmental impact. By addressing these aspects, our work contributes to responsible AI adoption in EDA, ensuring that automation enhances creativity and inclusivity in circuit design rather than reinforcing existing barriers.





\bibliography{main}
\bibliographystyle{icml2025}

\newpage
\appendix
\onecolumn

\section{Appendix: Baseline Model Details}
\label{sec:baseline}

We compare \textsc{AutoCircuit-RL} against various baseline models, including both LLM-based and non-LLM approaches, to assess its efficiency and performance in circuit topology synthesis.

\begin{itemize}
    \item \textbf{GPT-Neo-2.7} \cite{black2021gpt} \footnote{\url{https://huggingface.co/EleutherAI/gpt-neo-2.7B}}: A 2.7B parameter transformer-based model developed by EleutherAI, following GPT-3's architecture. It was trained on 420B tokens over 400,000 steps using masked autoregressive modeling with cross-entropy loss.
    
    \item \textbf{StableLM-3B-4E1T} \footnote{\url{https://huggingface.co/stabilityai/stablelm-3b-4e1t}}: A 3B parameter decoder-only model pre-trained on 1T tokens over 4 epochs from diverse English and code datasets. Referred to as StableLM in our work.

    \item \textbf{Llama-3-8B} \footnote{\url{https://huggingface.co/meta-llama/Meta-Llama-3-8B}}: An 8B parameter model from the Meta Llama 3 family, incorporating supervised fine-tuning (SFT) and reinforcement learning with human feedback (RLHF) to enhance alignment with human preferences.

    \item \textbf{MPT-7B} \footnote{\url{https://huggingface.co/mosaicml/mpt-7b}}: A 7B parameter decoder-style transformer trained on 1T tokens of English text and code. It employs a modified transformer architecture optimized by MosaicML for efficient training and inference.

    \item \textbf{\textsc{CircuitSynth-Gumbel}}: A variant of \cite{vijayaraghavan2024circuitsynth} incorporating:
    \begin{enumerate}
        \item Training a circuit validity and efficiency classifier to estimate the probability of a generated circuit being valid.
        \item Fine-tuning an LLM to generate circuit topologies.
        \item Refining outputs using the classifier while enforcing circuit validity and efficiency constraints.
    \end{enumerate}
    The training objective combines standard negative log-likelihood loss ($LL_{LM}$) with circuit validity and efficiency loss. Since LLMs operate in discrete spaces, we employ Gumbel-softmax \cite{jang2016categorical} for continuous relaxation, enabling gradient-based optimization.

    \item \textbf{GraphVAE} \cite{simonovsky2018graphvae}: A non-LLM baseline for circuit generation. The encoder consists of two graph convolutional layers (32 and 64 channels) with identity connections, batch normalization, and ReLU activation, followed by a fully connected (FC) layer producing a 256-dimensional latent representation. The decoder has three FC layers (256, 512, 1024 channels) with batch normalization and ReLU, followed by a parallel triplet of FC layers to output graph tensors. Training was performed for 50-75 epochs using Adam (learning rate $1e^{-3}$, $\beta_1=0.5$). A sentence transformer \cite{reimers2019sentence} processes natural language prompts, followed by an FC layer to align its representation with the encoder output size.
\end{itemize}


\begin{table*}[h]
\centering
\small
\scriptsize
\begin{tabular}{@{}lll@{}}
\toprule
Prompt Type   & Prompt     & Sample Circuit     \\ \midrule
Component Constraint      & \begin{tabular}[c]{@{}l@{}}Generate a 4-component circuit with n-type MOSFETs: FET-A-0, FET-A-1 and FET-A-2; \\ and p-type MOSFET: FET-B-0; representing different nodes:\\ {[}`FET-A-2',`FET-B-0',`FET-A-1', `FET-A-0'{]}\end{tabular}                                                        & \begin{tabular}[c]{@{}l@{}}{[}{[}`FET-A-2', `5', `0'{]},\\   {[}`FET-B-0', `5', `OUT'{]},\\   {[}`FET-A-1', `0', `IN'{]},\\   {[}`FET-A-0', `5', `OUT'{]}{]}\end{tabular}             \\
Output Voltage Constraint     & \begin{tabular}[c]{@{}l@{}}Generate a 4-component circuit with capacitors: capacitor-0 and capacitor-1; n-type \\ MOSFET: FET-A-0; and p-type MOSFET: FET-B-0 representing different nodes: \\{[}`capacitor-1',`FET-A-0',`capacitor-0', `FET-B-0'{]}  with Vout less than 1.5V when Vin \\ equals 2V\end{tabular} & \begin{tabular}[c]{@{}l@{}}{[}{[}`capacitor-1', `IN', `10'{]},\\   {[}`FET-A-0', `OUT', `IN'{]},\\   {[}`capacitor-0', `IN', `0'{]},\\   {[}`FET-B-0', `OUT', `10'{]}{]}\end{tabular} \\
Efficiency Constraint & \begin{tabular}[c]{@{}l@{}}Generate a 4-component circuit with inductor: inductor-0; n-type MOSFET: FET-A-0;\\ p-type MOSFET: FET-B-0; and capacitor: capacitor-0; representing different nodes:\\ {[}`inductor-0',`FET-A-0',`FET-B-0', `capacitor-0'{]} with efficiency greater than 0.6\end{tabular}             & \begin{tabular}[c]{@{}l@{}}{[}{[}`inductor-0', `6', `OUT'{]},\\   {[}`FET-A-0', `OUT', `6'{]},\\   {[}`FET-B-0', `IN', `6'{]},\\   {[}`capacitor-0', `OUT', `0'{]}{]}\end{tabular}    \\ \bottomrule
\end{tabular}
\caption{Sample prompts of each prompt type and their corresponding sample circuit topologies.}
\label{tab:example}
\end{table*}




\section{Appendix: Implementation Details}
\label{app:implement}
We provide the implementation details of our experiments conducted with the official PyTorch v2.2.0 release binary package, compiled with CUDA 11.8, utilizing NVIDIA V100 GPUs with 32 GB of memory.
\begin{itemize}
\item Two-Phase Training: This setup applies to both instruction tuning and RL-refinement phases. We plot the training data on a Vout-Efficiency graph (an example for 5-component circuits is shown in Figure \ref{fig:groups}) and categorize the data into four main groups: (a) Group 1 with low efficiency (efficiency lower than 0.05), (b) Group 2 with moderate efficiency (efficiency between 0.05 and 0.7), (c) Group 3 with high efficiency but the output voltage shows minimal difference from input voltage, and (d) Group 4 with optimal Vout and efficiency. To optimize the training process, we apply a weighted sampling strategy in each batch, giving the highest priority to Group 4, which represents the optimal conditions, and the lowest priority to Group 1, which has the least efficiency. The sampling weights are allocated as follows: 0.1 for Group 1, 0.25 for Group 2, 0.25 for Group 3, and 0.4 for Group 4. This approach ensures that batches are enriched with data from Group 4, allowing the model to better learn from the most critical and desirable conditions.

Training is conducted over 4-6 epochs using shuffled data from the training split, with model checkpoints saved based on the best performance on the validation split. To manage memory efficiently, we employ gradient checkpointing. We use the AdamW optimizer \cite{loshchilov2017decoupled}, setting beta parameters to 0.9 and 0.95, and an epsilon value of 1.0e-8. The learning rate is set to 0.95e-5, and the seed is fixed at 42 to ensure reproducibility. During training, we assess performance by evaluating a subset of 100 sample generations, using consistent evaluation settings. If the performance in the current epoch exceeds that of the previous one, we save the checkpoint.



\begin{figure}[h]
\centering
\includegraphics[scale=0.14]{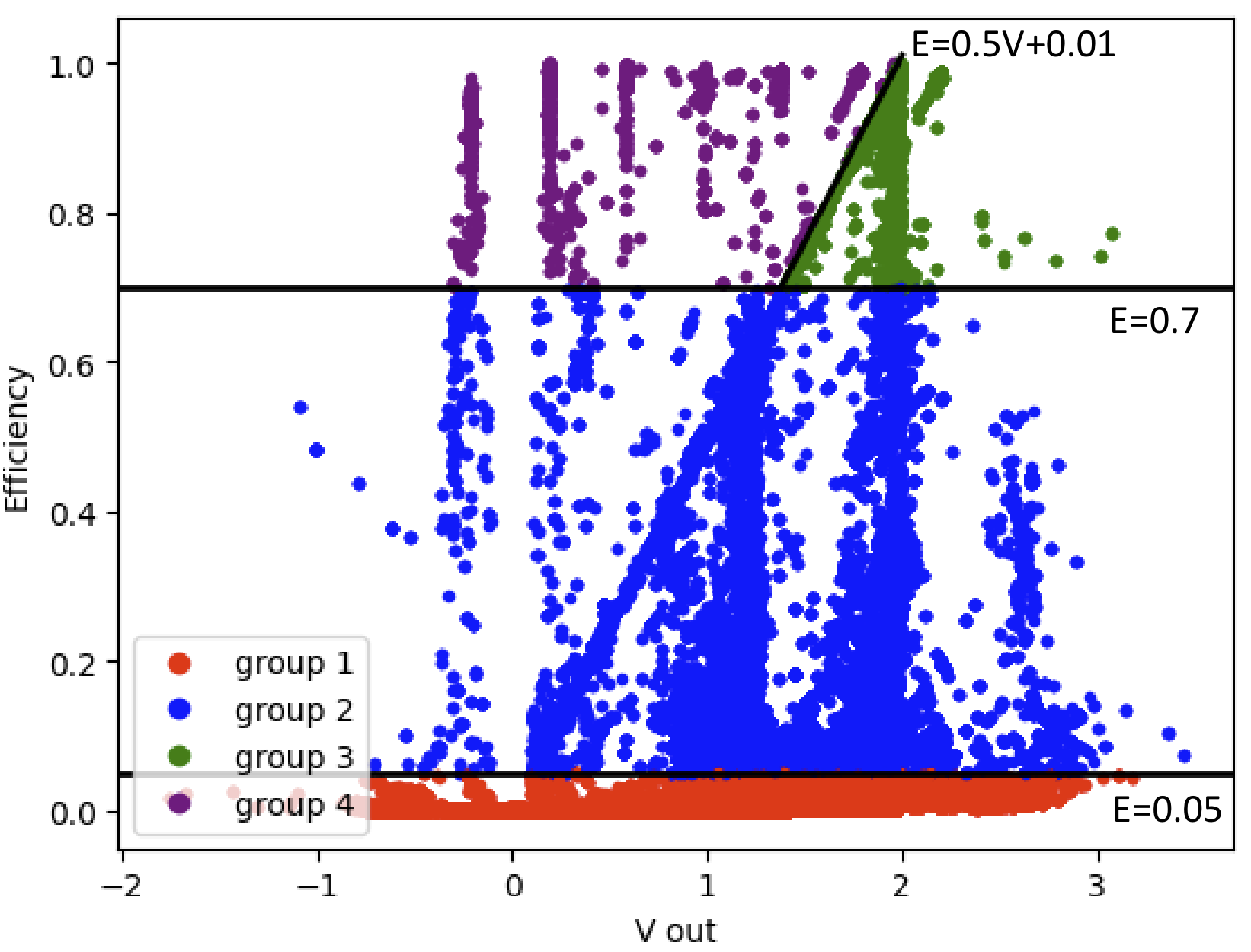}

\caption{Vout-Efficiency graph of 5-component circuits in the training data.}
\label{fig:groups}
\end{figure} 

\item Iterative Adaptation: The iterative adaptation, explained in Section \ref{sec:ia}, is performed over 3–5 iterations depending on the base model used. Each iteration involves 2–4 epochs of RL refinement and uses a dataset of 10,000 high-quality circuit samples obtained via nucleus sampling. This process improves performance progressively with each stage. The final adapted model is directly used for inference without further tuning.

\item Inference: We assess all models by generating 1,000 unique sample circuit topologies from each, using a combination of nucleus sampling and top-k sampling techniques. These generated topologies are then analyzed using validity, efficiency, and output voltage estimators to identify the designs that are both valid and efficient, meeting the specified design criteria.
\end{itemize}

\section{Error Analysis}
\label{sec:error-analysis}

\subsection{Validity Errors}

We begin our error analysis by examining one common failure mode encountered in generated circuit topologies: validity errors, as identified by SPICE simulation. A netlist is considered invalid if it violates essential electrical constraints such as correct node referencing, grounding, or connectivity, which causes simulation failures.

Despite the strong overall performance of our model, a small subset of generated netlists exhibit such validity issues. These typically stem from improper or inconsistent node assignments rather than errors in component selection or overall topology. A representative example of an invalid netlist generated by the model is:

\begin{verbatim}
[['FET-B-1', 'IN', '6'], 
 ['FET-A-0', '0', 'IN'], 
 ['FET-B-0', 'OUT', '7'], 
 ['inductor-0', '6', '7']]
\end{verbatim}

This netlist fails validation primarily due to ambiguous or incorrect node labeling, which disrupts the circuit's connectivity and prevents successful simulation. However, with minimal adjustments, specifically refining the node assignments to properly reference outputs and grounds, the netlist can be made valid:

\begin{verbatim}
[['FET-B-1', 'IN', '6'], 
 ['FET-A-0', '0', 'IN'], 
 ['FET-B-0', 'OUT', '0'], 
 ['inductor-0', '6', 'OUT']]
\end{verbatim}

This minor correction preserves the original component arrangement and topology while resolving the node reference ambiguities, enabling successful simulation. This pattern indicates that the model effectively learns appropriate component placements and connectivity patterns but occasionally struggles with precise node labeling. These validity errors are therefore close to valid designs and could be mitigated with improved node management during generation.

\subsection{Efficiency Constraint Errors}

In addition to validity, our generated circuits are evaluated against performance constraints such as minimum required efficiency, as specified in the generation prompt. These constraints are verified post-hoc using SPICE simulation. While the majority of the model outputs achieve or closely approximate the requested efficiency thresholds, a few circuits fall short. Consider the following example, where the prompt required an efficiency greater than 0.7:

\begin{verbatim}
[['inductor-0', 'IN', '6'], 
 ['FET-B-0', 'OUT', '6'], 
 ['capacitor-0', '0', 'OUT'], 
 ['FET-A-0', 'OUT', 'IN']]
\end{verbatim}

This topology, under a generated duty cycle of 0.1, achieved a simulated efficiency of 0.625, missing the constraint by a relatively small margin. The component usage and connectivity suggest that the model has captured many of the structural features that contribute to efficiency, although precise performance can depend on subtle circuit-level interactions. Such cases demonstrate that the model generates solutions near the constraint boundary and could be improved further with targeted refinement.

However, there also exist a few outlier cases where the efficiency gap is more pronounced, such as outputs scoring well below the target (for example, less than 0.3 when the required efficiency was greater than 0.7). These larger discrepancies are typically associated with difficult trade-offs between duty cycle, output voltage, and interactions among other components. In these cases, the model may prioritize satisfying voltage or topological structure over efficiency, reflecting the inherent challenge of simultaneously satisfying multiple constraints. Nevertheless, the proximity of many failing cases to the desired efficiency threshold, along with the model's ability to produce performance-aware topologies, supports the conclusion that these errors arise from nuanced constraint balancing rather than fundamental model shortcomings.

\section{Runtime Complexity Analysis}
\label{app:runtime}
A key strength of \textsc{AutoCircuit-RL} lies in its computational efficiency, particularly when compared with traditional search-based topology synthesis methods. These traditional methods, such as genetic algorithms and tree-based search~\cite{ICCAD21, Topo_Tree}, often rely on hundreds of SPICE simulations per design iteration. As a result, they typically require several minutes (hundreds of seconds) to produce a single valid circuit design, especially when adapting to new specifications or exploring diverse topologies.

In contrast, our method significantly reduces synthesis time by leveraging a two-phase RL approach. The LLM quickly generates initial candidate topologies, while the RL-based refinement performs iterative optimization using a learned reward model that captures circuit validity, efficiency, and output voltage. On average, \textsc{AutoCircuit-RL} generates a complete and optimized circuit in approximately 2–3.5 seconds using two NVIDIA V100 GPUs, offering over 50x improvement in design time over traditional approaches.

To assess how the choice of language model affects runtime, we evaluated \textsc{AutoCircuit-RL} with two LLMs: MPT-7B and LLaMA-3 8B. As expected, model size influences generation latency. The MPT-7B variant achieves faster runtimes (2.4–3.5 seconds for 4–10 component circuits), whereas the LLaMA-3 8B variant incurs slightly higher runtimes (2.8–5 seconds), reflecting the increased computational overhead of the larger model. Despite this, both configurations maintain runtimes well below traditional baselines, making them viable for real-time or interactive design applications.

\begin{figure}[h]
\centering
\includegraphics[width=0.5\linewidth]{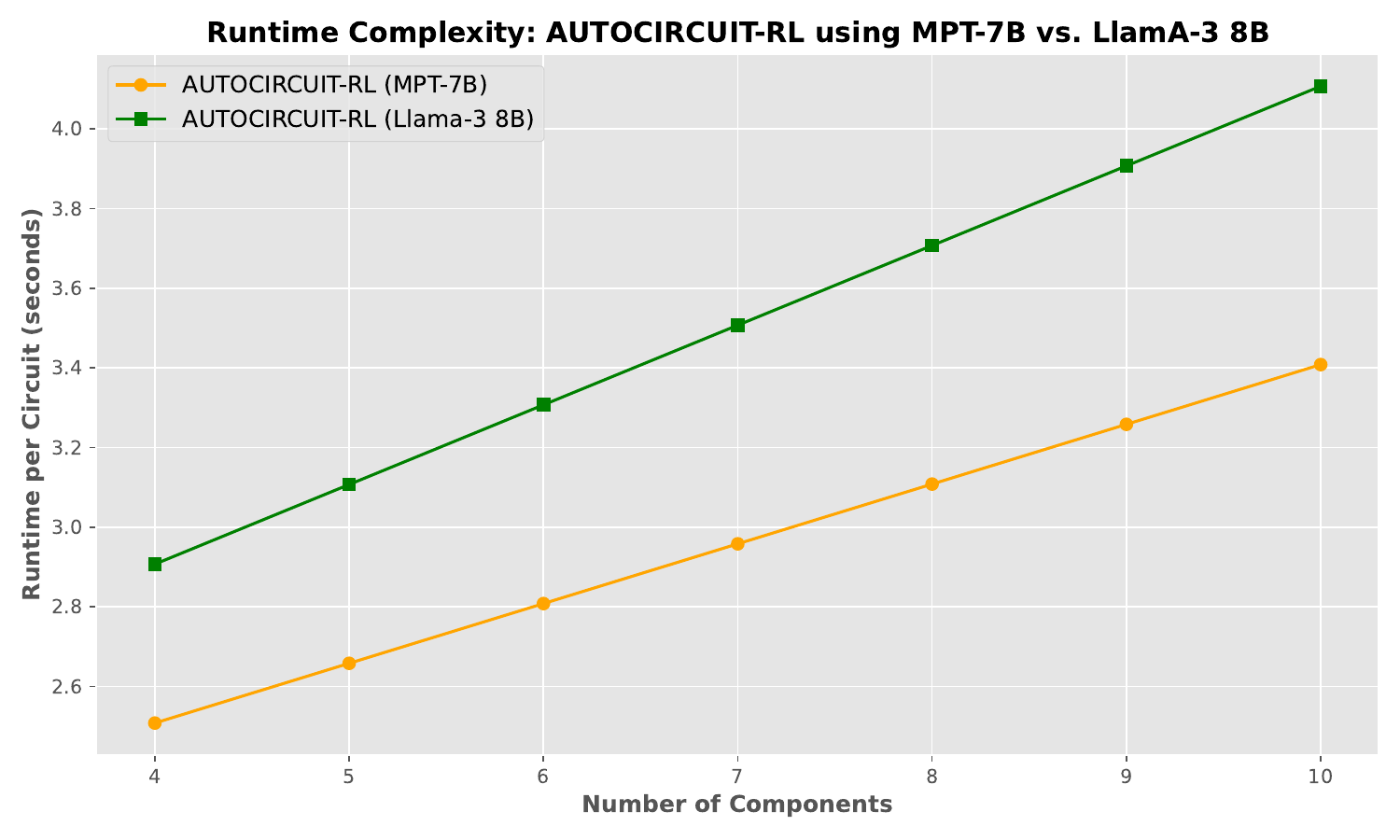} 
\caption{Runtime Complexity of \textsc{AutoCircuit-RL} Using MPT-7B vs. LLaMA-3 8B. The runtime per circuit increases with the number of components but remains significantly faster than traditional search-based methods. MPT-7B offers lower latency due to its smaller model size, while LLaMA-3 8B provides marginally higher runtimes due to increased model capacity.}
\label{fig:runtime}
\end{figure}

These results demonstrate that \textsc{AutoCircuit-RL} achieves an effective balance between computational cost and output quality, offering a scalable and practical alternative to traditional methods in analog circuit synthesis.

\end{document}